\documentclass[runningheads]{llncs}

\usepackage{eccv}

\usepackage{eccvabbrv}

\usepackage{graphicx}
\usepackage{booktabs}
\usepackage{amssymb}
\usepackage{marvosym}
\usepackage{float}
\usepackage{longtable}
\usepackage{multirow}

\usepackage{amsthm,amsmath,amssymb}
\usepackage[accsupp]{axessibility}  

\usepackage{hyperref}

\usepackage{orcidlink}

\makeatletter
\def\thanks#1{\protected@xdef\@thanks{\@thanks
		\protect\footnotetext{#1}}}
\makeatother

\begin{document}

\title{Pro2SAM: Mask Prompt to SAM with Grid Points for Weakly Supervised Object Localization} 

\titlerunning{Pro2SAM}

\author{Xi Yang\inst{1}\orcidlink{0000-0002-5791-3674} \and
Songsong Duan\inst{1}\orcidlink{0000-0003-2983-4044}\textsuperscript{\Letter}\thanks{\textsuperscript{\Letter}Corresponding author} \and
Nannan Wang\inst{1}\orcidlink{0000-0002-4695-6134}  \and
Xinbo Gao\inst{2}\orcidlink{0000-0003-1443-0776}}

\authorrunning{Xi Yang, Songsong Duan, et al.}

\institute{Xidian University \and
Chongqing University of Posts and Telecommunications
\\
 \email{yangx@xidian.edu.cn, duanss@stu.xidian.edu.cn, nnwang@xidian.edu.cn, gaoxb@cqupt.edu.cn}}

\maketitle

\begin{abstract}
  Weakly Supervised Object Localization (WSOL), which aims to localize objects by only using image-level labels, has attracted much attention because of its low annotation cost in real applications. Current studies focus on the Class Activation Map (CAM) of CNN and the self-attention map of transformer to identify the region of objects. However, both CAM and self-attention maps can not learn pixel-level fine-grained information on the foreground objects, which hinders the further advance of WSOL. To address this problem, we initiatively leverage the capability of zero-shot generalization and fine-grained segmentation in Segment Anything Model (SAM) to boost the activation of integral object regions. Further, to alleviate the semantic ambiguity issue accrued in single point prompt-based SAM, we propose an innovative mask prompt to SAM (Pro2SAM) network with grid points for WSOL task. First, we devise a Global Token Transformer (GTFormer) to generate a coarse-grained foreground map as a flexible mask prompt, where the GTFormer jointly embeds patch tokens and novel global tokens to learn foreground semantics. Secondly, we deliver grid points as dense prompts into SAM to maximize the probability of foreground mask, which avoids the lack of objects caused by a single point/box prompt. Finally, we propose a pixel-level similarity metric to come true the mask matching from mask prompt to SAM, where the mask with the highest score is viewed as the final localization map. Experiments show that the proposed Pro2SAM achieves state-of-the-art performance on both CUB-200-2011 and ILSVRC, with 84.03\% and 66.85\% Top-1 Loc, respectively.
\end{abstract}

\section{Introduction}
\label{sec:intro}

Over the past decade, computer vision has relied on deep learning and a wealth of labeled data to achieve huge successes. However, fine-grained manual annotations are time-consuming and expensive, especially for large amounts of data, which hinders the development of computer vision methods in real-world scenarios. To relieve this issue, weakly supervised learning that uses incomplete annotations as training data has drawn increasingly extensive attention in various computer vision tasks, such as object localization, semantic segmentation, visual grounding, and referring expression comprehension. Weakly Supervised Object Localization (WSOL), one of the most eye-catching research directions, aims to locate the foreground objects by using only image-level supervision without bounding box annotations. WSOL alleviates massive efforts in obtaining fine annotations and holds widespread significance in dataset creation and related visual tasks \cite{ref-48}.

\begin{figure*}[t]
	\centering\includegraphics[width=0.98\textwidth,height=6cm]{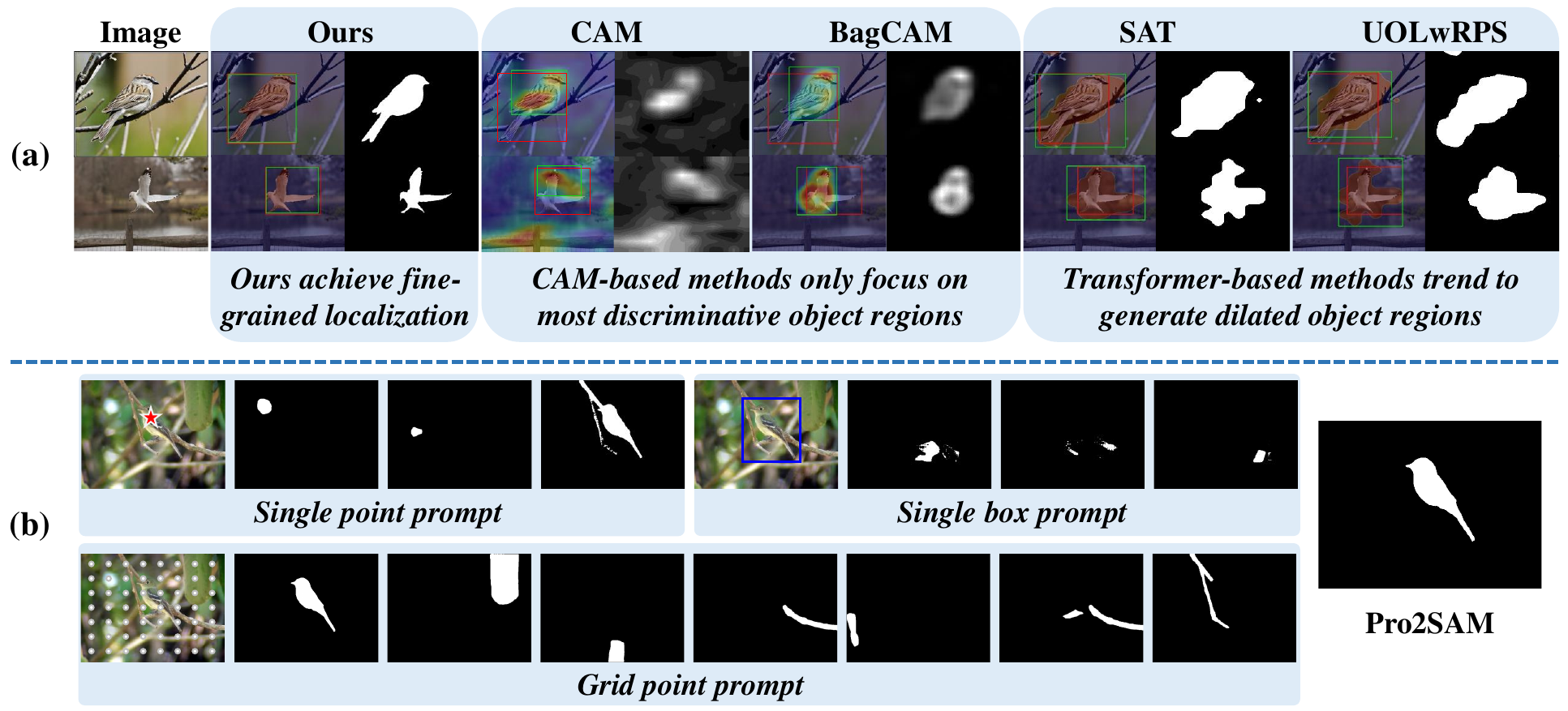}
	\caption{(a) Comparison of Ours, CAM-based (CAM \cite{ref-1} and BagCAM \cite{ref-42}), and transformer-based methods (SAT \cite{ref-6} and UOLwRPS \cite{ref-46}). The \textcolor{red}{red} and \textcolor{green}{green} boxes are GT bounding boxes and predicted bounding boxes, and box prompt, respectively. (b) Ambiguity of SAM, \textcolor{blue}{blue} bounding box is box prompt and red star is point prompt.}
	\label{Fig.1}	
	\vspace{-2.0em}	 	
\end{figure*}

As a pioneer of WSOL, the work of CAM \cite{ref-1} discovered that the learned class weights in class-specific fully connected layers can inherently reflect the localization information by activating deep feature representation. Unfortunately, CAM tends to focus only on the most discriminative object regions, which is insufficient for accurate object localization. To address this problem, many alternative strategies are proposed to remedy CAM's salient region activation problem, including erasing salient regions \cite{ref-2, ref-3}, context information \cite{ref-4}, and class-agnostic localization maps \cite{ref-5, ref-6}. Although these works improve the localization performance of WSOL, partial region activation of the CAM still exists and needs to be addressed urgently.

Recently, transformers \cite{ref-7, ref-9} have been introduced to computer vision and have succeeded vastly. The surprising self-attention mechanism captures long-range feature dependency, which implicitly represents the semantic correlations of patch tokens, thus the attention map upon patch tokens is pivotal for object localization in the transformer-based WSOL methods. TS-CAM \cite{ref-8} was the first work to introduce the transformer to the WSOL task by using a self-attention map to locate foreground objects. Further, SCM \cite{ref-10} and LCTR \cite{ref-11} enhanced the local continuity between the patch and its neighboring patches via the inherent spatial coherence of the object. With the transformer and its variants, partial region activation of CAM-based methods can be solved to some extent. As shown in Fig. 1 (a), the activation regions of transformer-based methods cover most of the object areas, which CAM cannot achieve.

Transformer-based WSOL methods employ self-attention maps to locate foreground objects via long-range dependence. Due to the transformer's expertise in learning global semantic information, there is a deficiency in perceiving object boundaries. Therefore, transformer-based methods often result in dilated activation regions, erroneously identifying the background surrounding the object as part of the object itself. As shown in Fig. 1 (a), SAT \cite{ref-6} and UOLwRPS \cite{ref-46} significantly activate the partial background, resulting in imprecise localization maps. We think the self-attention maps of the transformers are not an optimal solution for accurate object localization, especially when evaluating high IOU thresholds (\textit{e.g.}, 0.7 and 0.9). In brief, CAM-based and transformer methods can not capture fine-grained information, resulting in coarse localization results and hindering the development of accurate object localization.

To address this problem, we initiatively leverage the capability of zero-shot generalization and fine-grained segmentation in the Segment Anything Model (SAM) to improve the localization accuracy of WSOL models. SAM has robust zero-shot capability and segments objects with fine-grained boundaries, which can facilitate precise object localization. Unfortunately, SAM aims to segment potential objects and generate masks for each instance via the box/point prompts, including foreground and background, which leads to semantic ambiguity in SAM, as shown in Fig. 1 (b). To alleviate this issue, we propose a mask matching solution from the mask prompt to SAM (named Pro2SAM) with grid points for precise WSOL. First, we design a Global Token Transformer (GTFormer), which joint embedding patch tokens and novel global tokens into the self-attention to learn similarity dependency from global tokens to patch tokens and produces the localization map efficiently. Then, we send preset grid points into SAM to extract a mask gallery of images. As shown in Fig. 1 (b), the grid point prompts can improve the probability of a foreground mask, avoiding the lack of a foreground mask caused by a single point/box prompt. Finally, we employ pixel-level similarity as an evaluation metric to achieve mask matching between the mask prompt generated by GTFormer and the mask gallery of SAM, where the mask with the highest score is viewed as the final localization map.

1) We propose a novel Global Token Transformer (GTFormer) to learn global semantic representation via joint embedding patch tokens and global tokens. Specifically, GTFormer can model the mutual dependency of global and local token features to produce localization maps.

2) We first introduce SAM into the WSOL to leverage its zero-shot capability and fine-grained segmentation. Additionally, we propose a mask matching solution from the mask prompt generated by GTFormer to SAM for selecting the anticipated mask as the localization result, thereby addressing SAM's semantic ambiguity issue.

3) Extensive experiments show that Pro2SAM outperforms SOTA methods by a large margin on multiple benchmarks. Surprisingly, our model generated fine-grained localization maps and performs excellently even under extreme settings (\textit{e.g.}, 48.43\% and 45.96\% IOU90 localization accuracy on CUB-200-2011 and ILSVRC datasets).

\section{Related Work}

\subsection{Weakly Supervised Object Localization}

The WSOL aims to localize objects by solely image-level labels. The pioneer work CAM \cite{ref-1} demonstrated the effectiveness of localizing objects using deep features with fully connected weights from CNNs trained initially for classification networks. Despite its simplicity, CAM-based methods suffer from limited prominent regions, which cannot discover objects completely. 

To address this issue, many strategies were proposed to expand the activation, including adversarial erasing \cite{ref-2, ref-3}, discrepancy learning \cite{ref-12}, online localization refinement \cite{ref-13, ref-14}, task decoupling \cite{ref-15, ref-16}, and attention regularization \cite{ref-6, ref-9}. For example, adversarial erasing \cite{ref-2} online removed significantly activated regions within feature maps to learn the missed object parts. Spatial discrepancy learning leveraged adversarial classifiers to enlarge object areas. Through task decoupling, PSOL \cite{ref-16} partitioned the WSOL pipeline into two parts, \textit{e.g.}, class-agnostic object localization and object classification, to learn class-agnostic maps.

Recently, because of the advantage of capturing long-range dependencies, the transformer has been introduced into the field of WSOL. Compared to the CNN-based CAM methods, the transformer-based methods employed self-attention maps to localize relatively complete object regions. Dosovitskiy \textit{et al.} \cite{ref-7} demonstrated that a pure transformer performs exceptionally well on image classification tasks when applied directly to sequences of image patches. In WSOL, transformer-based models have also shown promising results. For instance, Gao \textit{et al.} \cite{ref-18} combined semantic-aware tokens with the semantic-agnostic attention map to find objects. Chen \textit{et al.} \cite{ref-11} highlighted local details of global representations using learnable kernels and cross-patch information guided by the class-token attention map. Gupta \textit{et al.} \cite{ref-20} improved localization maps by incorporating a patch-based attention dropout layer into the transformer attention blocks. The recent surge in attention towards pre-trained large models has led to some interesting attempts to incorporate them into the WSOL, yielding surprising results, such as CLIP \cite{ref-25} was used in \cite{ref-49}.

\subsection{Foundation Models}
Pre-trained Foundation Models (PFMs) have gained significant popularity because of their exceptional zero-shot capability in addressing unseen data distributions and tasks, alongside their remarkable transferability in learned representations. PFMs have been extensively explored in Natural Language Processing (NLP), leading to the development of milestone models, such as BERT \cite{ref-21}, BART \cite{ref-22}, and the GPT series \cite{ref-23, ref-24}. Similarly, the exploration of PFMs for visual-language studies has resulted in excellent models such as CLIP \cite{ref-25}. 

Segment Anything Model (SAM) \cite{ref-26} was recently introduced as the first foundation model for visual segmentation, pre-trained on over 1.1 billion masks. This model allowed users to interact by inputting various geometric prompts, demonstrating strong zero-shot capability across conventional image domains. Its potential in numerous applications represents a significant paradigm shift in both academic and industry fields, such as medical image processing \cite{ref-28, ref-30}, 3D vision \cite{ref-31, ref-33} , in-painting \cite{ref-32}, object tracking \cite{ref-34, ref-35}, and remote sensing \cite{ref-27, ref-29}.

For instance, the HQ-SAM \cite{ref-36} introduced a novel approach by freezing the entire pre-trained SAM and only prompt-tuning the mask decoder. It introduces a new mask head, enabling to generation of the original SAM mask and adaptive mask simultaneously. UV-SAM \cite{ref-29} adapted SAM to urban village segmentation via mixed prompts, including mask, bounding box, and image representations, then fed into SAM for fine-grained boundary identification. This paper is the first work introducing the SAM to WSOL via generated mask prompts.

\section{Methodology}

\subsection{ Overall Architecture}

As shown in Fig. 2, Pro2SAM comprises three parts: mask prompt generation stage, SAM assist stage, and mask matching stage. Consider the lack of fine-grained information for existing models, including CAM-based methods and transformer-based methods, we leverage SAM's robust edge detection and mask segmentation capability to learn fine-grained localization information. Following earlier works \cite{ref-6, ref-9}, we design a novel transformer-based localization model to predict a coarse localization map in the Mask Prompt Generation Stage. Then, we send the pre-set grid point into the prompt encoder of SAM to generate all fine-grained masks of given images in the SAM Assist Stage. Finally, we calculate the similarity scores of the coarse mask prompt generated by the GTFormer and fine-grained masks from SAM, and then select the one with the highest score as a final localization map.

\begin{figure*}[t]
	\centering\includegraphics[width=0.98\textwidth,height=7cm]{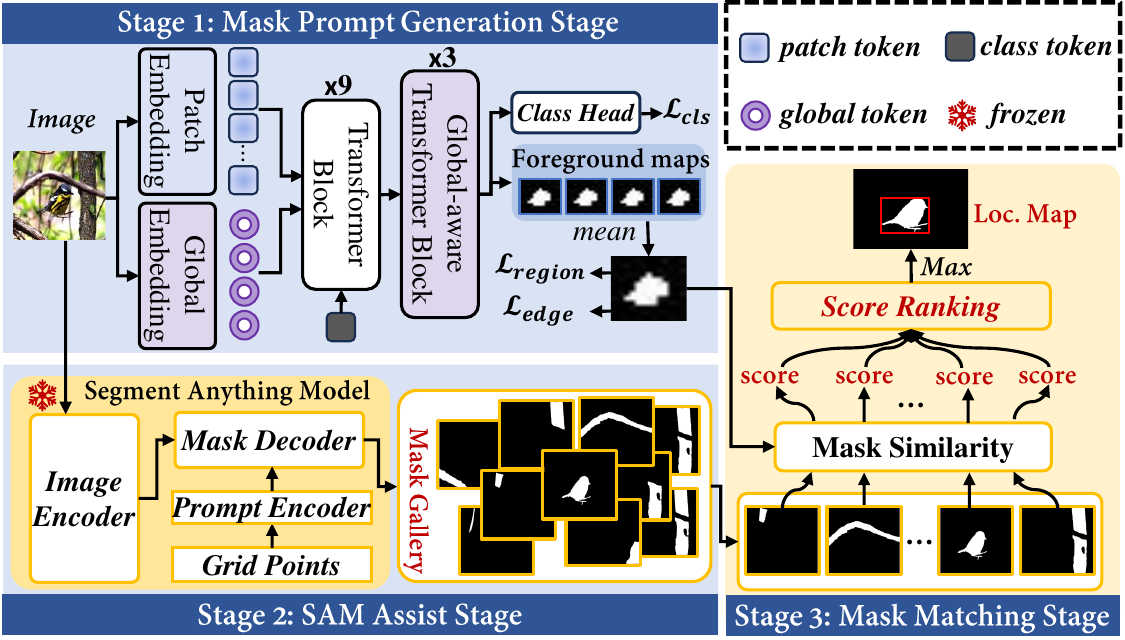}
	\caption{Overview of our proposed Pro2SAM framework, which comprises three stages: Stage 1) Mask Prompt Generation stage, we propose a Global Token Transformer (GTFormer) to predict coarse foreground maps via only image-level labels; Stage 2) SAM Assist Stage, we input the grid points of the image into SAM as prompts to generate all masks of input images; Stage 3) Mask Matching Stage, we evaluate the similarity scores between the coarse mask from GTFormer and fine-grained masks from SAM to select a perfect mask as a localization map.}
	\label{Fig.2}	
	\vspace{-2.0em}	 	
\end{figure*}

Given an input image represented by $I \in \mathbb{R}^{3 \times h \times w}$, WSOL aims to generate the localization map $M_{Loc} \in \mathbb{R}^{1 \times h \times w}$ by a localization model learned only with the image-level label $y \in \mathbb{R}^{1 \times C}$, where \textit{h} and \textit{w} are the height and width of the image, and \textit{C} is number of classes. The proposed GTFormer splits image \textit{I} into a sequence of non-overlapping patches and four global patches, as shown in Fig. 3. Then the divided patches and global patches are flattened and transformed into patch tokens $x_p \in \mathbb{R}^{N \times D}$ and global tokens $x_g \in \mathbb{R}^{4 \times D}$ by linear projection, where the $N = \frac{h}{P} \times \frac{w}{P}$ and $D$ is the sequence length and channels of patch tokens, $P$ is the size of patches. For simplicity, we omit the description of the batch size B. After grouping the class token $x_c$ and the global tokens $x_g$ with patch tokens $x_p$, this token sequence $\mathcal{F}^{0} \in \mathbb{R}^{(N+5)\times D}$ is sent into stacked transformer blocks and global aware transformer blocks for subsequent representation learning. After GTFormer, we can get a coarse mask $M_{b}$ as a foreground map. With grid points, SAM can generate all potential masks $M_{SAM}^{L}$, where $L$ is the number of masks generated by SAM for a given image $I$. Note that SAM uses a complex post-processing operation to exclude some noisy masks, which results in a different number of masks for each image. Finally, we select a well-matched mask from $M_{SAM}^{L}$ as localization result $M_{Loc}$ via similarity score.

\subsection{Global Token Transformer}

\begin{figure*}[t]
	\centering\includegraphics[width=0.78\textwidth,height=5cm]{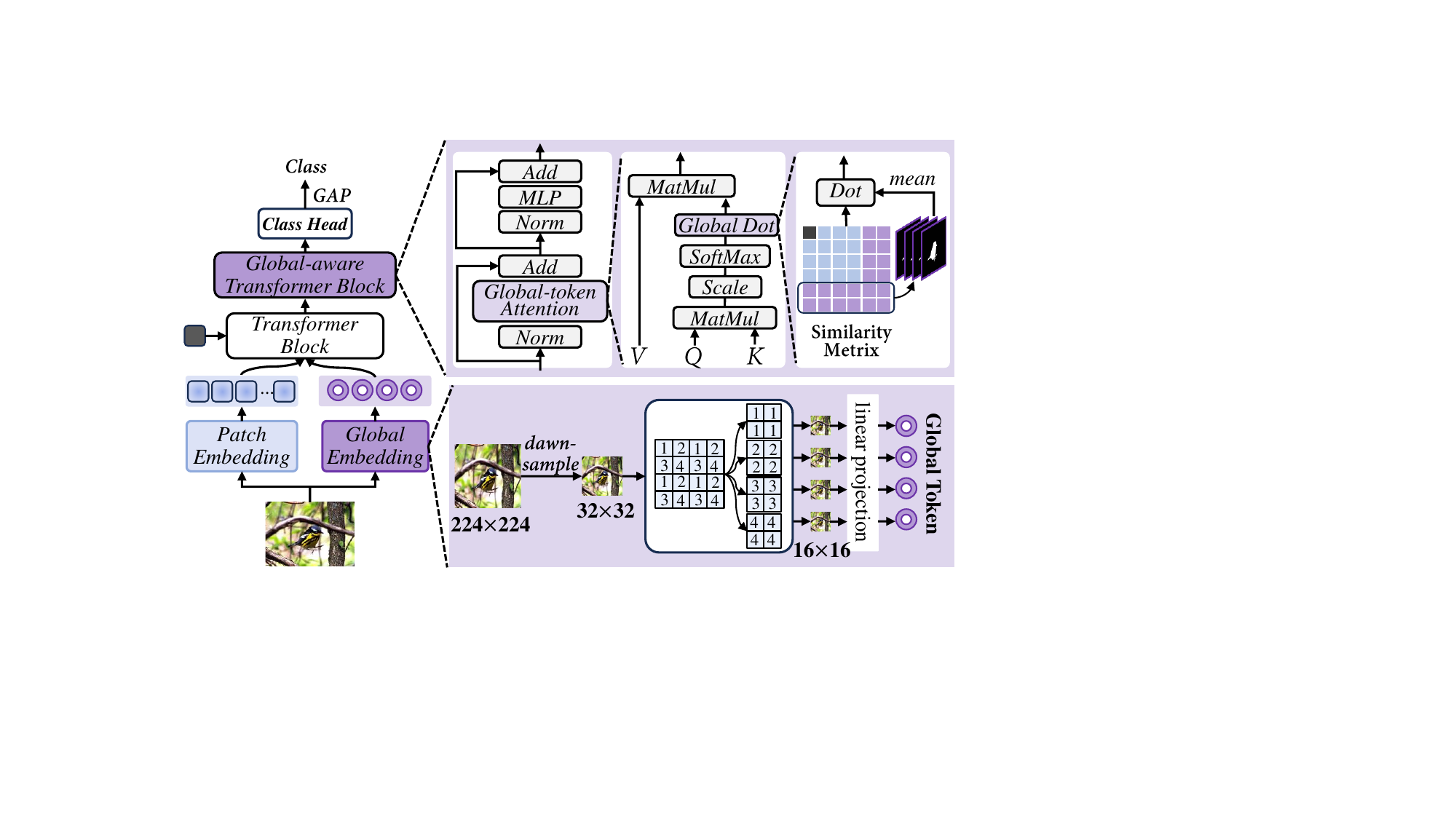}
	\caption{Overview of our proposed GTFormer, which inherits the architecture of transformer blocks for ViT \cite{ref-7, ref-17} and adds novel global tokens and global-aware transformer blocks. Furthermore, we propose a global embedding layer to learn global tokens.}
	\label{Fig.3}	
	\vspace{-2.0em}	 	
\end{figure*}

Recently, some works have used self-attention in transformers for long-range dependency to re-active semantic regions, aiming to avoid discriminative region activation in traditional CAM. However, the transformer splits input images into disconnected patches for subsequent representation learning and dependency modeling, neglecting the inherent spatial coherence of the object \cite{ref-10}. It diffuses the semantic-aware regions far from the object boundary, dilating localization maps significantly. Fig. 1 (a) presents several visual comparisons to show the dilated activation regions with coarse boundary. To address this issue, we propose global tokens to learn global spatial coherence from original images, and then heighten the semantic connection of separable patches via global-aware transformer blocks.

\textbf{\textit{Global Embedding.}} As shown in Fig. 3, an input image is resized to $224 \times 224$, then down-sampled to $32 \times 32$. To reserve the complete global spatial structure of the image, the conventional max/average pooling is ill-suited. Inspired by patch merging in Swin Transformer \cite{ref-37}, we employ a $4 \times 4$ window to scan the original image with stride set to 4, then the same position pixels in the $4 \times 4$ window are reconstructed into the global patches. After linear projection, the global patches are translated into global tokens. The total processing can be defined as:
	%

\begin{align}
	\left\{  
	\begin{aligned}
		x_g^{1}&=f_{pj}(Win_{(1,1)}(I_{dw})), \\
		x_g^{2}&=f_{pj}(Win_{(1,2)}(I_{dw})), \\
		x_g^{3}&=f_{pj}(Win_{(2,1)}(I_{dw})), \\
		x_g^{4}&=f_{pj}(Win_{(2,2)}(I_{dw})),
	\end{aligned}
	\right.
\end{align}
where $I_{dw}$ is the down-sampled original image, and $f_{pj}$ is the linear projection. $Win_{(1,1)}$, $Win_{(1,2)}$, $Win_{(2,1)}$, and $Win_{(2,2)}$ indicate the global sample in different positions from $I_{dw}$ via a fixed window shift. $x_g = \{x_g^{1}, x_g^{2}, x_g^{3}, x_g^{4}\}$ is the sequence of global tokens, which is stacked with patch tokens and class token in self-attention block for mutual relational modeling. 

\textbf{\textit{Global-aware Transformer Block.}} The proposed global-aware transformer block inherits the typical transformer block structure. Thus, GTFormer differs from the classic transformer by the alternative attention layers and token sequences. For an input token sequence $\mathcal{F}^{l-1}$, GTFormer learn a localization-aware token sequence $\mathcal{F}^{l}$ via the global-token attention in a interaction manner, which can be defined as:
\begin{align}
	\mathcal{F}^{l-1} &= Cat(x_g, x_p, x_c), \\
	\mathcal{X}^{l} & = LayerNorm(\mathcal{F}^{l-1}), \\
	\mathcal{K}^{l} & = \mathcal{F}^{l-1} + GTA(\mathcal{X}^{l}), \\
	\mathcal{F}^{l} & = \mathcal{K}^{l} + MLP(LayerNorm(\mathcal{K}^{l})),
\end{align}
where the $Cat$ is the token connection operation, and $GTA$ is the global-token attention in the global-aware transformer block, the detail of global-token attention is described in Fig. 3.

The proposed global-token attention is based on a self-attention module and aims at converting the global-aware token $x_g$ into a localization map $M_b$ by querying different patches. Meanwhile, the generated $M_b$ is applied to the cross-attention calculation as a dot product, thus obtaining localization information from the image-level labels. Specifically, global-token attention first linearly projects the token sequence $\mathcal{X}^{l}$ onto the query matrix Q, key matrix K, and value matrix V, where $Q, K, V \in \mathbb{R}^{(N+5)\times D}$. 

Then the query vector corresponding to the global token $\{Q_{glb}^{i}\}_{i=1}^{4} \in \mathbb{R}^{1 \times D}$ is selected from Q and used to query K to get the query results for each token. After applying sigmoid activation, the similarity map is transformed into foreground probability with a 0 to 1 interval distribution. The foreground probability map between the spatial query vector and the key matrix in the $i$-th layer is calculated as follows: 
\begin{equation}
	M_{glb}^{i} = Sigmoid(\dfrac{Q_{glb}^{i} \cdot K^T}{\sqrt{W}}) \in \mathbb{R}^{1\times (N+5)},
\end{equation}
where $i$ means different global tokens, and $\sqrt{W}$ is used as a scaling factor, like \cite{ref-7}. The generated $\{M_{glb}^{i}\}_{i=1}^{4}$ serves as a visual cue to emphasize the object region and is combined in the calculation of attention with the form of the dot product, which can be formulated as:
\begin{align}
	M_{glb} & = \dfrac{1}{4}\sum_{i=1}^{4}M_{glb}^{i}, \\
	GTA(\mathcal{X}^{l}) &= Softmax(\dfrac{QK^T}{\sqrt{W}})*M_{glb}\cdot V,
\end{align}
where $*$ denotes element-wise multiplication, and $\cdot$ is dot product.

By cropping and reshaping, the part of the $\{M_{glb}^{i}\}_{i=1}^{4}$ to the patch tokens can be transformed into a localization map $\{M_{l}^{i}\}_{i=1}^{4} \in \mathbb{R}^{\frac{h}{P} \times \frac{w}{P}}$. After obtaining  $\{M_{l}^{i}\}_{i=1}^{4}$ learned from different global tokens, we take their averages as the final coarse localization map $M_b$ for subsequent mask matching. To optimize GTFomer, besides the classification loss, we use edge loss \cite{ref-6} and region loss \cite{ref-6} to enhance the localization ability of the network. Therefore, the total loss function can be formulated as follows:
\begin{equation}
	\mathcal{L} = \mathcal{L}_{cls} + \mu*(\mathcal{L}_{edge}^{M_b} + \mathcal{L}_{region}^{M_b}) + \lambda*\sum_{i=1}^{4}(\mathcal{L}_{edge}^{M_{l}^{i}} + \mathcal{L}_{region}^{M_l^{i}})
\end{equation}
where the $\mathcal{L}_{cls}$ is the cross-entropy classification loss, followed with \cite{ref-8}, and $\mu$ and $\lambda$ are weight factors to control optimizing strategy.

\subsection{Mask Matching via SAM}
The segment anything model (SAM) is the first foundation model for general image segmentation. It has achieved impressive results on various natural image segmentation tasks. In this paper, we introduce SAM into the WSOL via mask matching. Recently, some works \cite{ref-18, ref-29, ref-27, ref-33, ref-34} substantiated SAM has two strong advantages: 1) powerful zero-shot capability; and 2) fine-grained segmentation capability. Therefore, SAM can generate masks with finer boundaries for unseen images. The two characteristics can effectively improve the localization accuracy of the WOSL task. 

However, the masks generated by SAM do not have any semantic information, and it does not know which category the generated mask belongs to, which confuses the WSOL methods. To address the problem, we propose a mask matching scheme to select a well-match mask as the final localization results. Different from some works \cite{ref-29, ref-30}, which fine-tune SAM on domain-specific datasets and employ a single point or box prompt to generate masks of object regions, our method adopt preset grid point prompts to erase semantic ambiguity of SAM. Furthermore, we only train GTFormer without the need to fine-tune SAM, which greatly reduces training time and computational power, the SAM and mask matching only are used for inference. Specifically, we fed the grid point prompts into the prompt encoder of SAM to generate a mask gallery, which can be defined as:
\begin{equation}
	M_{SAM}^{L} = SAM(I, [point_1, ..., point_G]),
\end{equation}
where the $M_{SAM}^{L}$ is obtained mask gallery and $L$ is the length of $M_{SAM}^{L}$. $[point_1, ..., point_G]$ are grid point prompts, and G is the number of points. 

After SAM, we can get a mask gallery $M_{SAM}^{L}$, which contains all object masks, whether foreground or background. WSOL aims to discover a suitable mask form $M_{SAM}^{L}$, which belongs to foreground objects. So, we compute the pixel-level similarity between $M_{SAM}^{L}$ and coarse localization map $M_b$ generated by GTFormer, which can be described as:
\begin{equation}
	Score_i = \frac{\sum_{H=1}^{h/p}\sum_{W=1}^{w/p}f_{and}(M_{SAM}^{i}, M_b)}{\sum_{H=1}^{h/p}\sum_{W=1}^{w/p}f_{or}(M_{SAM}^{i}, M_b)}
\end{equation}
where $f_{and}$ and $f_{or}$ denote logical or and logical and operations. $Score_i$ indicates the mask similarity between $M_{SAM}^{i}$ and $M_b$, then we can get the similarity score list of $M_{SAM}^{L}$. Note that $M_{SAM}^{i}$ is resized to keep the same resolution with $ M_b$. Finally, we select the mask with the highest score as the final localization map.

\begin{table*}[t]
	\large
	\renewcommand\arraystretch{2.5}
	\centering
	\fontsize{8}{8}
	\selectfont
	\caption{\label{comparison} Quantitative comparison of localization and classification accuracy with state-of-the-art methods on CUB-200-2011 and ILSVRC datasets. The best results are highlighted in \textbf{bold}, and second are \underline{underlined}.}
	\label{tab:distortion_type}
	\setlength{\tabcolsep}{2pt} 
	\renewcommand\arraystretch{1.2}
	\resizebox*{\textwidth}{!}{
		\large
		\begin{tabular}{lccc|ccc|ccc}
			\toprule[2pt]
			
			\multirow{2}{*}{\textbf{Method}} &
			\multirow{2}{*}{\textbf{Publication}} &
			\multirow{2}{*}{\textbf{Loc. Back.}} &
			\multirow{2}{*}{\textbf{Cls. Back.}} &
			
			\multicolumn{3}{|c|}{\textbf{CUB-200-2011 }} &
			\multicolumn{3}{c}{\textbf{ILSVRC }} \cr
			
			& & & & Top-1 Loc & Top-5 Loc & GT-Known & Top-1 Loc & Top-5 Loc & GT-Known \cr \hline  \hline

			CAM \cite{ref-1} &	CVPR16 & \multicolumn{2}{c|}{GoolLeNet} &		36.10 &	50.66 &	56.00 &	42.80 &	54.90 &	59.00 \cr
			
			GCNet \cite{ref-40}  &	ECCV20 & \multicolumn{2}{c|}{Inception V3}	&	58.58 &	71.00 &	75.30 &	49.06 &	58.09&	-      \cr
			ORNet \cite{ref-13}  &	ICCV21 & \multicolumn{2}{c|}{VGG16}		    &	67.73 &	80.77 &	86.20 &	52.05 &	63.94&	68.27  \cr
			TS-CAM \cite{ref-8}  &  ICCV21 & \multicolumn{2}{c|}{Deit-S}		&	71.30 &	83.80 &	87.70 &	53.4  &	64.30&	67.60  \cr
			BAS \cite{ref-14}    &	CVPR22 & \multicolumn{2}{c|}{ResNet50}	    &	77.25 &	90.08 &	95.13 &	57.18 &	68.44&	71.77  \cr
			CREAM \cite{ref-41}  &	CVPR22 & \multicolumn{2}{c|}{Inception V3}  &	71.76 &	83.87 &	90.43 &	56.07 &	66.19&	69.03  \cr
			BagCAMs \cite{ref-42}&  ECCV22 & \multicolumn{2}{c|}{ResNet50}	    &	69.67 &	  -   &	94.01 &	44.24 &	  -  &	72.08  \cr
			SCM \cite{ref-10}    &	ECCV22 & \multicolumn{2}{c|}{Deit-S}		&	76.40 &	91.60 &	96.60 &	56.10 &	66.40&	68.80  \cr
			LCAR \cite{ref-44}   &	AAAI23 & \multicolumn{2}{c|}{Vit-S/16}      &	77.40 &	  -   &	95.90 &	57.10 &	 -   &	70.70  \cr
			SAT \cite{ref-6}     &	ICCV23 & \multicolumn{2}{c|}{Deit-S}		&	80.43 &	93.58 &	\underline{98.19} &	59.94 &	\underline{70.31}&	73.08  \cr
			CATR \cite{ref-9}    &	ICCV23 & \multicolumn{2}{c|}{Deit-S}		&	79.62 &	92.08 &	94.74 &	56.9  &	66.64&	69.25  \cr \hline 
			Ours-GTFormer        &This Work& \multicolumn{2}{c|}{Deit-S}	    &	80.48 &	\textbf{93.84} & \textbf{98.48} &	\underline{60.2}  &	70.00&	\underline{73.17}  \cr
			
			\hline  \hline
			
			PSOL  \cite{ref-16}   &	CVPR20   &	DenseNet161&	EfficientNet-B7&    74.97 &	89.12 &	93.01 &	55.31 &	64.18 &	66.28\cr
			SPOL  \cite{ref-43}   &	CVPR21   &	ResNet50   &	EfficientNet-B7&	80.12 &	93.44 &	96.46 &	56.40 &	67.15 &	69.02\cr
			C2AM \cite{ref-15}    &	CVPR22   &	DenseNet161&	EfficientNet-B7&	81.76 &	91.11 & 92.88 &	59.56 &	67.05 &	68.53\cr
			TokenCut \cite{ref-45}&	CVPR22   &	DINO       &	EfficientNet-B7&	72.90 &	  -   &	91.80 &	52.30 &	  -	  &  65.40\cr
			UOLwRPS	\cite{ref-46} & ICCV23   &	MoCo V2	   &    EfficientNet-B7&	79.57 &	92.60 &	96.67 &	55.60 &	66.05 &	69.10\cr
			LocLoc \cite{ref-47}  &	ACM MM23 &	ViT-S      &	DINO+CrabCut   &	\textbf{84.40} &	93.30 &	98.10 &	57.60 &	67.20 &	70.00\cr
			\hline
			
			Ours-Pro2SAM  &	This Work &	Deit-S+SAM & EfficientNet-B7 &	\underline{84.03} &	\underline{93.69} &	95.67 	& \textbf{66.85}	& \textbf{75.90}	& \textbf{77.79} \cr
			
			\bottomrule[2pt]
			
	\end{tabular}}
\end{table*}

\section{Experiments}

\subsection{Experiment Settings}
\indent \textit{\textbf{Datasets.}} We evaluate Pro2SAM on two commonly used benchmarks, CUB-200-2011 \cite{ref-19} and ILSVRC \cite{ref-38}. CUB-200-2011 is a fine-grained dataset with 200 bird species, which consists of 5,994 images for training and 5,794 images for testing. ILSVRC is a large-scale dataset, which contains about 1.2 million images with 1,000 categories for training and 50,000 images for validation. We use only the image-level labels for training, while for evaluation, we incorporate both the image-level annotations and bounding box annotations.

\noindent \textbf{\textit{Evaluation Metrics.}} Following prior studies \cite{ref-14, ref-15}, We evaluate the performance by the commonly used metric Top-1/Top-5 localization accuracy (Top-1/Top-5 Loc.), maximal box accuracy (MaxBoxAccV2) \cite{ref-39}, and GT-known localization (GT-Known ) accuracy as our evaluation metrics. For GT-Known metric, a bounding box prediction is positive if its Intersection-over-Union (IoU) with at least one of the ground truth boxes is over 50\%. Top-1/Top-5 Loc. metric is correct when the ground-truth class belongs to the Top-1/Top-5 prediction categories and GT-Known is correct.

\noindent \textit{\textbf{Implementation Details.}} The Deit-S \cite{ref-17} pre-trained on ImageNet is used as the backbone of our GTFormer.  The input images are resized to 256 × 256 and then randomly cropped to 224 × 224. We use an AdamW optimizer with $\epsilon$=1e-8, $\beta_1$=0.9, $\beta_2$=0.99, and weight decay of 5e-4, to train our network on 4 NVIDIA RTX 4090 GPUs.  The training process lasts 30 epochs with a batch size of 128 for CUB-200-2011 and 5 epochs with a batch size of 256 for ILSVRC. Note that Only the GTFormer is trained on CUB-200-2011 and ILSVRC, where the $\mu$ and $\lambda$ of total loss are set to 1.0 and 0.5. The SAM only is used in the inference with ViT-H model to generate mask gallery.

\begin{table*}[t]
	\large
	\renewcommand\arraystretch{2.5}
	\centering
	\fontsize{8}{8}
	\selectfont
	\caption{\label{comparison} Comparison of MaxBoxAccV2 scores in terms of IOU50, IOU70 and IOU90 on the CUB-200-2011 and ILSVRC datasets for evaluating fine-grained localization. The best results of WSOL methods are highlighted in \textbf{bold}.}
	\label{tab:distortion_type}
	\setlength{\tabcolsep}{6pt} 
	\renewcommand\arraystretch{1.2}
	\resizebox*{\textwidth}{!}{
		\large
		\begin{tabular}{lc|cccc|cccc}
			\toprule[2pt]
			
			\multirow{2}{*}{\textbf{Method}} &
			\multirow{2}{*}{\textbf{Publication}} &
			\multicolumn{4}{|c|}{\textbf{CUB-200-2011 }} &
			\multicolumn{4}{c}{\textbf{ILSVRC }} \cr
			
			& & IOU50 & IOU70 & IOU90 & Mean & IOU50 & IOU70 & IOU90 & Mean \cr \hline  \hline 
			
			CAM 	\cite{ref-1}  &  CVPR16&	73.30 &	19.90 &	  -	  &   -	  & 65.70 &	41.60 &	  -	  &    -  \cr
			TS-CAM	\cite{ref-8}  &  ICCV21&	87.70 &	50.10 &	  -	  &   -	  & 66.10 &	44.60 &	  -	  &    -  \cr
			SPOL	\cite{ref-43} &  CVPR21&	93.22 &	45.72 &	 0.40 &	46.45 &	62.81 &	44.59 &	11.89 &	39.76 \cr
			BAS	    \cite{ref-14} &  CVPR22&	93.62 &	66.86 &	 6.33 &	55.60 &	69.68 &	47.57 &	 5.68 &	40.98 \cr
			BagCAM	\cite{ref-42} &  ECCV22&	87.57 &	27.80 &	 0.60 &	38.66 &	68.30 &	46.59 &	 3.95 &	39.61 \cr
			TokenCut\cite{ref-45} &  CVPR22&	92.87 &	75.21 &	25.96 &	64.68 &	  -	  &	  -	  &   -   &   -   \cr
			SAT	    \cite{ref-6}  &  ICCV23&	94.88 &	58.64 &	 2.01 &	51.84 &	69.77 &	51.96 &	15.14 &	45.62 \cr
			CATR	\cite{ref-9}  &  ICCV23&	78.82 &	39.44 &	 1.44 &	39.90 &	33.51 &	10.96 &	 0.34 &	14.94 \cr
			UOLwRPS	\cite{ref-46} &  ICCV23&	96.19 &	63.93 &	 4.35 &	54.82 &	63.19 &	39.84 &	 5.00 &	36.01 \cr
			
			LocLoc	\cite{ref-47}  & ACM MM23 &	95.61 &	78.00 &	15.85 &	63.15 &	67.57 &	46.49 &	14.68 &	42.91 \cr
			\hline
			Ours-GTFormer          & This Work&	\textbf{97.40} &	77.04 &	 8.70 &	61.05 &	70.45 &	50.71 &	12.36 &	44.51 \cr
			Ours-Pro2SAM           & This Work&	95.67 &	\textbf{86.43} &	\textbf{48.43} &	\textbf{76.84} &	\textbf{77.79} &	\textbf{66.93} &	\textbf{45.94} &	\textbf{63.55} \cr
			\bottomrule[2pt]
			
	\end{tabular}}
\end{table*}

\subsection{Comparison with State-of-the-Arts}
\textbf{Quantitative Comparison.} To demonstrate the effectiveness of the Pro2SAM, we compare it against previous methods on CUB-200-2011 and ILSVRC in Table. 1. Following previous works \cite{ref-15, ref-43, ref-46}, we conduct two comparisons with incorporated localization methods \cite{ref-1,ref-6,ref-9, ref-8, ref-10, ref-13, ref-14, ref-40, ref-41, ref-42, ref-44} and class-agnostic localization methods \cite{ref-15, ref-16,ref-43, ref-45, ref-46, ref-47}. For incorporated localization methods, GTFormer outperforms all compared methods using various backbones on the CUB-200-2011 dataset and gets the best localization results in terms of Top-1 Loc and GT-Known metrics compared with other SOTA methods on the ILSVRC dataset. Besides, GTFormer designed a novel transformer structure to generate localization maps, which obtained 75.86\% and 24.02\% performance gains in GT-Known metric compared with pioneer work CAM \cite{ref-1} on the CUB-200-2011 and ILSVRC datasets, respectively. For class-agnostic methods, Pro2SAM get the best performances on the ILSVRC dataset and the second results on CUB-200-2011. Nevertheless, Pro2SAM still get good results with a slight gap compared with the LocLoc \cite{ref-47}.

\begin{figure*}[t]
	\centering\includegraphics[width=0.93\textwidth,height=7cm]{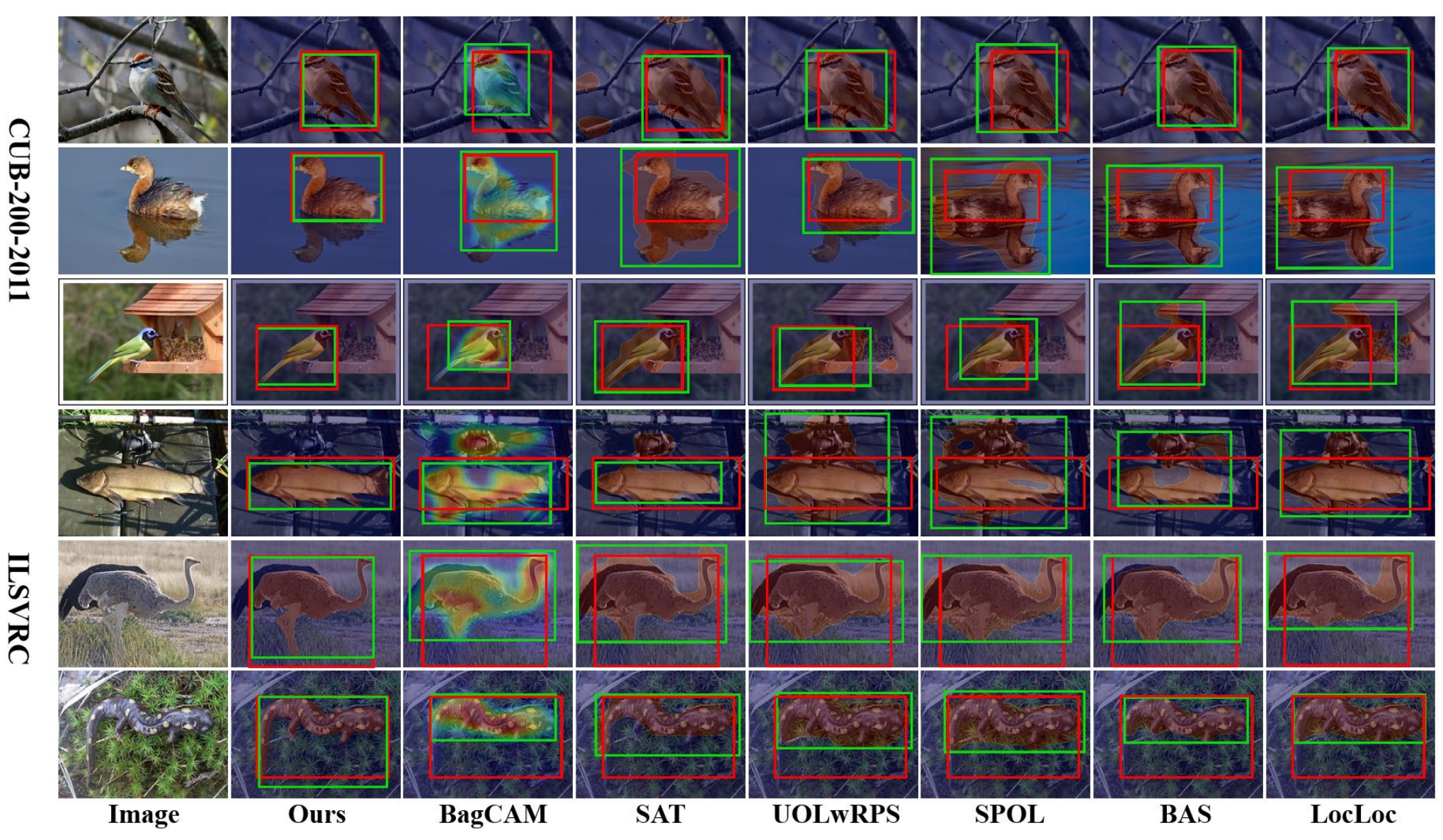}
	\caption{Visualization comparison. The ground-truth bounding boxes are in 
		\textcolor{red}{red}, and the predicted bounding boxes are in 
		\textcolor{green}{green}.}
	\label{Fig.4}	
\end{figure*}

\begin{table}[htbp]
	\centering
	\begin{minipage}{0.48\textwidth}
		\centering
		\footnotesize 
		\begin{tabular}{cccc}
			\toprule
			Method  & Top-1 Loc & GT-Known \\
			\midrule
			SAT \cite{ref-6}    & 59.9 & 73.0 \\
			Self-Att. Map  & 43.0 & 54.7 \\
			w/o-GTA    & 51.1 & 63.3 \\
			GTFormer   & 60.2 & 73.2 \\
			\bottomrule
		\end{tabular}
		\caption{Ablation studies of GTFormer components. w/o-GTA is the variant by removing GTA from Ours}
		\label{table:gtformer}
	\end{minipage}\hfill 
	\begin{minipage}{0.48\textwidth}
		\centering
		\footnotesize 
		\begin{tabular}{cccc}
			\toprule
			$\mu$:$\lambda$ & Top-1 Loc & GT-Known \\
			\midrule
			1:0 	& 59.6 & 72.1 \\
			0:1  	& 59.4 & 71.2 \\
			1:1 	& 43.6 & 54.7 \\
			0.5:1  	& 59.3 & 72.4 \\
			1:0.5  	& 60.2 & 73.2 \\
			\bottomrule
		\end{tabular}
		\caption{Ablation studies for numbers of Global Token on ILSVRC dataset.}
		\label{table:sample}
	\end{minipage}
\end{table}

\textbf{Fine-grained Localization.} WSOL aims to find and locate the objects from the confused background. Therefore, it is very important for the WSOL task to judge the model by evaluating the quality of localization. Previous studies \cite{ref-1, ref-6, ref-46} usually used GT-known ($IOU \geq 0.5$) to evaluate localization quality. We think GT-Known is not suitable for fine-grained localization ($IOU \geq 0.7$ or $0.9$). So, we report MaxBoxAccV2 scores with IOU50, IOU70, and IOU90 of some methods \cite{ref-1,ref-6,ref-8,ref-9,ref-14,ref-42,ref-43,ref-45,ref-46} in Table 2. Pro2SAM improves IOU70 and IOU90 metrics by 14.92\% and 18.80\% compared with the second method TokenCut \cite{ref-45} on CUB-200-2011, and 28.81\% and 203.43\% compared with the second method SAT \cite{ref-45} on ILSVRC. Obviously, Pro2SAM significantly exceeds these methods for fine-grained localization.

\textbf{Visual Comparison.} Besides quantitative comparisons, we also present visual comparisons of localization maps with Pro2SAM, BagCAM, SAT, UOLwRPS, SPOL, BAS, and LocLoc on CUB-200-2011 and ILSVRC datasets in Fig. 4. Compared with the BagCAM, which is an advanced CAM method, the localization results of our Pro2SAM cover more complete object regions with tighter bounding boxes. Compared with SAT, UOLwRPS, which usually generates dilated localization maps via the tailored transformer, our Pro2SAM generate fine-grained localization maps with sharp edges. Besides, our model effectively address some complex scenes. For example, on the second line of Fig. 4, our method locate the duck without an inverted reflection in the water, the localization maps of other methods almost cover the inverted reflection.

\subsection{Ablation Study}
In this subsection, we implement a series of ablation experiments. All experiments are conducted on ILSVRC, a universal dataset, to increase the generalizability of the experimental results.

\textbf{GTFormer components.} We study the function of self-attention and global token to prove its effectiveness, the compared results are reported in Table 3. We find that the self-attention map is far below our global token. Besides, we also prove the effectiveness of global-token attention by comparing GTFormer-w/o-GTA and GTformer in Table 3.

%
%
%
%

\begin{table*}[t]
	\large
	\renewcommand\arraystretch{2.5}
	\centering
	\fontsize{8}{8}
	\selectfont
	\caption{\label{comparison} Ablation studies for numbers of Global Token on ILSVRC dataset. The results of Ours are highlighted in \textbf{bold}.}
	\label{tab:distortion_type}
	\setlength{\tabcolsep}{6pt} 
	\renewcommand\arraystretch{1.2}
	\resizebox*{0.9\textwidth}{!}{
		\large
		\begin{tabular}{ccccc}
			\toprule[1pt]
			
			\multirow{1}{*}{\textbf{Numbers of Global Token}} &
			\multirow{1}{*}{\textbf{Parameters (M)}} &
			\multirow{1}{*}{\textbf{Top-1 Loc}} &
			\multirow{1}{*}{\textbf{Top-5 Loc}} &
			\multirow{1}{*}{\textbf{GT-Known}}  \cr \hline 
			
			1	   & 25.42 &   59.4        &  69.6            & 72.3          \cr
			\textbf{4} & \textbf{26.31} & \textbf{60.2} &  \textbf{70.0 }  & \textbf{73.2} \cr
			16	   & 29.89 &  41.3         &  48.1            & 50.3          \cr	
			
			\bottomrule[1pt]
			
	\end{tabular}}
\end{table*}

\begin{table*}[t]
	
	\large
	\renewcommand\arraystretch{2.5}
	\centering
	\fontsize{8}{8}
	\selectfont
	\caption{\label{comparison} Ablation studies of mask Matching mechanism on ILSVRC dataset. The results of Ours are highlighted in \textbf{bold}.}
	\label{tab:distortion_type}
	\setlength{\tabcolsep}{6pt} 
	\renewcommand\arraystretch{1.2}
	\resizebox*{0.7\textwidth}{!}{
		\large
		\begin{tabular}{cccccc}
			\toprule[1pt]
			
			\multirow{1}{*}{\textbf{Strategies}} &
			\multirow{1}{*}{\textbf{IOU50}} &
			\multirow{1}{*}{\textbf{IOU60}} &
			\multirow{1}{*}{\textbf{IOU70}} &
			\multirow{1}{*}{\textbf{IOU80}} &
			\multirow{1}{*}{\textbf{IOU90}}  \cr \hline

			BBox Matching         &	 53.3 &	47.5 &	42.3 &	37.3 &	29.6    \cr
			Mask \& BBox Matching &	 67.3 &	62.2 &	56.8 &	51.7 &	40.3    \cr
			\textbf{Mask Matching}  &  \textbf{77.8} &	\textbf{72.6} &	\textbf{66.9} &	\textbf{60.0} &	\textbf{45.9}    \cr
			
			\bottomrule[1pt]
			
	\end{tabular}}
\end{table*}

\textbf{Hyperparameters $\mu$ and $\lambda$.} For the GTFormer, we employ two hyperparameters $\mu$ and $\lambda$ to control the parameter updating of GTFormer. In this subsection, we conduct experiments to confirm the values of $\mu$ and $\lambda$. As shown in Table 4, the GTFormer get the best results when $\mu$ and $\lambda$ are set to 1.0 and 0.5. Besides, we also conduct the same experiments on the CUB-200-2011, and set the values of $\mu$ and $\lambda$ to 1.0 and 0.5.

\textbf{Numbers of Global Token.} The global tokens, as a novel element for our GTFormer, generate similarity maps, which are used as localization maps. Therefore, the number of global tokens is a vital factor for localization quality. As shown in Table 5, we listed some results of variants with different numbers of global tokens. We find that GTFormer generates the best performance with 4 global tokens.

\textbf{Mask Matching Mechanism.} For Pro2SAM, we used mask matching to identify the best-suitable mask from the mask gallery generated by SAM as a final localization map. Furthermore, we formulate two alternative matching mechanisms by computing IOU scores of GT and predicted bounding boxes. As shown in Table 6, Mask Matching is a promising solution for Pro2SAM.

\subsection{Discussions}

\textbf{SAM \textit{vs} Transformer-based WSOL methods.} As we know, transformer-based WSOL methods usually outperform CAM-based methods via global long-range dependencies. However, Transformer-based methods inevitably produce dilation of object region. Our Pro2SAM introduces SAM \cite{ref-26} to address this issue. We count the IOU score with different thresholds to analyze and prove the fine-grained segmentation capability of SAM in Fig. 5, which indicates that Pro2SAM get more accurate segmentation compared with representative transformer-based methods, such as SAT \cite{ref-6} and UOLwRPS \cite{ref-46}. Besides, we also testify to the superiority of Pro2SAM by comparing it with an advanced CAM-based method BagCAM \cite{ref-42}.

\begin{figure*}[t]
	\centering\includegraphics[width=0.65\textwidth,height=2.75cm]{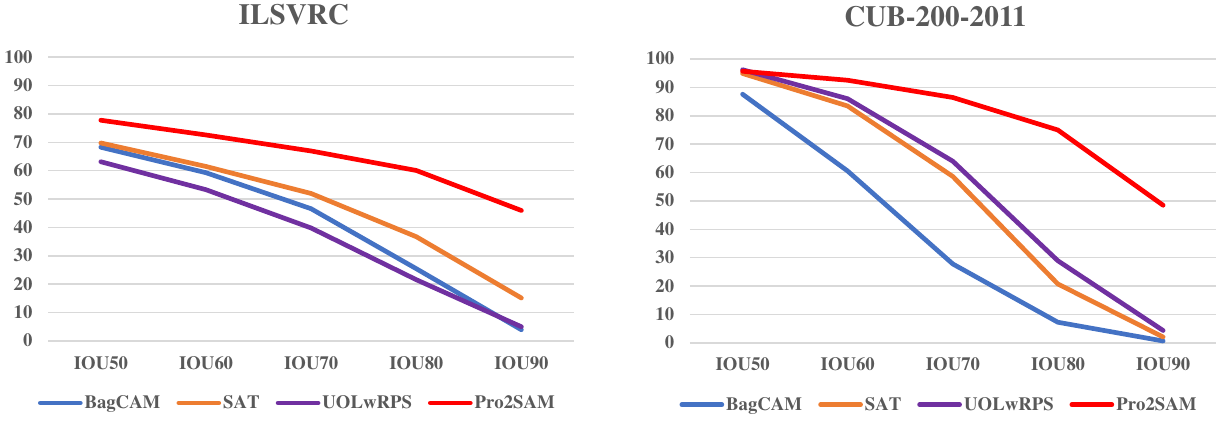}
	\caption{Accurate localization comparison of Pr2SAM and other SOTA methods with diverse IOU thresholds.}
	\label{Fig.5}	
\end{figure*}

\textbf{Grid points \textit{vs} Single box/point.} Prompt learning is a vital component in SAM. How to prompt in SAM is important for Pro2SAM. We analysis three prompt types: single point, single box, and grid points. We find single point/box prompt easily result in omission of object masks and noise masks, a simple sample is shown in Fig. 1 (b). Due to space constraints, we present detailed quantitative and qualitative comparisons of single point, single box, and grid points in \textit{Supplementary Material}.

\section{Conclusion}
This paper proposes to introduce the Segment Anything Model (SAM) to WSOL due to its powerful zero-shot generalization and fine-grained segmentation. Based on SAM, we propose a mask matching solution from generated mask prompt to SAM to address the semantic ambiguity of SAM. To this end, we design a Global Token Transformer (GTFormer) to learn a coarse localization map as mask prompt for the subsequent mask matching. The proposed GTFormer jointly embeds novel global tokens and patch tokens into the self-attention mechanism to learn similarity dependency from global tokens to patch tokens. Finally, we employ pixel-level similarity as evaluation scores to come true mask matching. Extensive experiments on ILSVRC and CUB-200-2011 datasets verify the effectiveness and efficiency of the proposed Pro2SAM that surpasses previous methods by a large margin.


\section*{Acknowledgements}
This work was supported in part by the National Natural Science Foundation of China under Grant 62372348, Grant 62441601, Grant U22A2096 and Grant 62221005; in part by the Key Research and Development Program of Shaanxi under Grant 2024GX-ZDCYL-02-10; in part by Shaanxi Outstanding Youth Science Fund Project under Grant 2023-JC-JQ-53; in part by the Innovation Collaboration Special Project of Science, Technology and Innovation Bureau of Shenzhen Municipality under Project CJGJZD20210408092603008; in part by the Fundamental Research Funds for the Central Universities under Grant QTZX24080 and Grant QTZX23042. 

%
%
\bibliographystyle{splncs04}
\bibliography{main}

\begin{thebibliography}{10}
\providecommand{\url}[1]{\texttt{#1}}
\providecommand{\urlprefix}{URL }
\providecommand{\doi}[1]{https://doi.org/#1}

\bibitem{ref-23}
Achiam, O., Adler, S., Agarwal, S.: Gpt-4 technical report  (2023)

\bibitem{ref-10}
Bai, H., Zhang, R., Wang, J., Wan, X.: Weakly supervised object localization
  via transformer with implicit spatial calibration. In: Proceedings of the
  ECCV (2022)

\bibitem{ref-47}
Cao, X., Zheng, X., Shen, Y., Li, K., Chen, J., Lu, Y., Tian, Y.: Locloc:
  Low-level cues and local-area guides for weakly supervised object
  localization. In: Proceedings of the ACM MM. pp. 5655--5664 (2023)

\bibitem{ref-31}
Cen, J., Zhou, Z., Fang, J., Shen, W., Xie, L., Jiang, D., Zhang, X., Tian, Q.,
  et~al.: Segment anything in 3d with nerfs. In: Proceedings of the NeurIPS.
  vol.~36 (2024)

\bibitem{ref-9}
Chen, Z., Ding, J., Cao, L., Shen, Y., Zhang, S., Jiang, G., Ji, R.:
  Category-aware allocation transformer for weakly supervised object
  localization. In: Proceedings of the ICCV. pp. 6643--6652 (2023)

\bibitem{ref-11}
Chen, Z., Wang, C., Wang, Y., Jiang, G., Shen, Y., Tai, Y., Wang, C., Zhang,
  W., Cao, L.: Lctr: On awakening the local continuity of transformer for
  weakly supervised object localization. In: Proceedings of the AAAI. p.
  410–418 (2022)

\bibitem{ref-39}
Choe, J., Oh, S.J., Lee, S., Chun, S., Akata, Z., Shim, H.: Evaluating weakly
  supervised object localization methods right. In: Proceedings of the CVPR.
  pp. 3133--3142 (2020)

\bibitem{ref-3}
Choe, J., Shim, H.: Attention-based dropout layer for weakly supervised object
  localization. In: Processing of the CVPR (2019)

\bibitem{ref-7}
Dosovitskiy, A., Beyer, L., Kolesnikov, A., Weissenborn, D., Houlsby, N.: An
  image is worth 16x16 words: Transformers for image recognition at scale. In:
  Proceedings of the ICLR (2021)

\bibitem{ref-18}
Feng, C.B., Lai, Q., Liu, K., Su, H., Vong, C.M.: Boosting few-shot semantic
  segmentation via segment anything model. arXiv preprint arXiv:2401.09826
  (2024)

\bibitem{ref-20}
Gupta, S., Lakhotia, S., Rawat, A., Tallamraju, R.: Vitol: Vision transformer
  for weakly supervised object localization. In: Proceedings of the CVPR. pp.
  4101--4110 (2022)

\bibitem{ref-28}
Huang, Y., Yang, X., Liu, L., Zhou, H., Chang, A., Zhou, X., Chen, R., Yu, J.,
  Chen, J., Chen, C., Liu, S., Chi, H., Hu, X., Yue, K., Li, L., Grau, V., Fan,
  D.P., Dong, F., Ni, D.: Segment anything model for medical images? Medical
  Image Analysis  \textbf{92},  103061 (2024)

\bibitem{ref-36}
Ke, L., Ye, M., Danelljan, M., Tai, Y.W., Tang, C.K., Yu, F., et~al.: Segment
  anything in high quality. In: Proceedings of the NeurIPS. vol.~36 (2024)

\bibitem{ref-21}
Kenton, J.D.M.W.C., Toutanova, L.K.: Bert: Pre-training of deep bidirectional
  transformers for language understanding. In: Proceedings of NAACL-HLT.
  vol.~1, p.~2 (2019)

\bibitem{ref-26}
Kirillov, A., Mintun, E., Ravi, N., Mao, H., Rolland, C., Gustafson, L., Xiao,
  T., Whitehead, S., Berg, A.C., Lo, W.Y., Dollár, P., Girshick, R.: Segment
  anything. In: Proceedings of the ICCV. pp. 3992--4003 (2023)

\bibitem{ref-22}
Lewis, M., Liu, Y., Goyal, N., Ghazvininejad, M., Mohamed, A., Levy, O.,
  Stoyanov, V., Zettlemoyer, L.: Bart: Denoising sequence-to-sequence
  pre-training for natural language generation, translation, and comprehension.
  In: Proceedings of the ACL (2020)

\bibitem{ref-37}
Liu, Z., Lin, Y., Cao, Y., Hu, H., Wei, Y., Zhang, Z., Lin, S., Guo, B.: Swin
  transformer: Hierarchical vision transformer using shifted windows. In:
  Proceedings of the ICCV. pp. 10012--10022 (2021)

\bibitem{ref-40}
Lu, W., Jia, X., Xie, W., Shen, L., Zhou, Y., Duan, J.: Geometry constrained
  weakly supervised object localization. In: Proceedings of the ECCV. pp.
  481--496 (2020)

\bibitem{ref-2}
Mai, J., Yang, M., Luo, W.: Erasing integrated learning: A simple yet effective
  approach for weakly supervised object localization. In: Processing of the
  CVPR (2020)

\bibitem{ref-30}
Mazurowski, M.A., Dong, H., Gu, H., Yang, J., Konz, N., Zhang, Y.: Segment
  anything model for medical image analysis: an experimental study. Medical
  Image Analysis  \textbf{89},  102918 (2023)

\bibitem{ref-5}
Meng, M., Zhang, T., Tian, Q., Zhang, Y., Wu, F.: Foreground activation maps
  for weakly supervised object localization. In: Processing of the ICCV (2021)

\bibitem{ref-4}
Pan, X., Gao, Y., Lin, Z., Tang, F., Dong, W., Yuan, H., Huang, F., Xu, C.:
  Unveiling the potential of structure preserving for weakly supervised object
  localization. In: Processing of the CVPR (2021)

\bibitem{ref-44}
Pan, Y., Yao, Y., Cao, Y., Chen, C., Lu, X.: Coarse2fine: local consistency
  aware re-prediction for weakly supervised object localization. In:
  Proceedings of the AAAI. vol.~37, pp. 2002--2010 (2023)

\bibitem{ref-25}
Radford, A., Kim, J.W., Hallacy, C., Ramesh, A., Goh, G., Agarwal, S., Sastry,
  G., Askell, A., Mishkin, P., Clark, J., Krueger, G., Sutskever, I.: Learning
  transferable visual models from natural language supervision. In: Proceedings
  of the ICML. vol.~139, pp. 8748--8763 (2021)

\bibitem{ref-35}
Raji{\v{c}}, F., Ke, L., Tai, Y.W., Tang, C.K., Danelljan, M., Yu, F.: Segment
  anything meets point tracking. arXiv preprint arXiv:2307.01197  (2023)

\bibitem{ref-38}
Russakovsky, O., Deng, J., Su, H., Krause, J., Satheesh, S., Ma, S., Huang, Z.,
  Karpathy, A., Khosla, A., Bernstein, M., et~al.: Imagenet large scale visual
  recognition challenge. International Journal of Computer Vision
  \textbf{115},  211--252 (2015)

\bibitem{ref-46}
Song, Y., Jang, S., Katabi, D., Son, J.: Unsupervised object localization with
  representer point selection. In: Proceedings of the ICCV. pp. 6534--6544
  (2023)

\bibitem{ref-24}
T, B., B, M., N, R., et~al: Language models are few-shot learners. In:
  Proceedings of the NeurIPS. pp. 1877--1901 (2020)

\bibitem{ref-17}
Touvron, H., Cord, M., Douze, M., Massa, F., Sablayrolles, A., J{\'e}gou, H.:
  Training data-efficient image transformers \& distillation through attention.
  In: Proceedings of the ICML. pp. 10347--10357. PMLR (2021)

\bibitem{ref-45}
Wang, Y., Shen, X., Hu, S.X., Yuan, Y., Crowley, J.L., Vaufreydaz, D.:
  Self-supervised transformers for unsupervised object discovery using
  normalized cut. In: Proceedings of the CVPR. pp. 14543--14553 (2022)

\bibitem{ref-43}
Wei, J., Wang, Q., Li, Z., Wang, S., Zhou, S.K., Cui, S.: Shallow feature
  matters for weakly supervised object localization. In: Proceedings of the
  CVPR. pp. 5993--6001 (2021)

\bibitem{ref-19}
Welinder, P., Branson, S., Mita, T., Wah, C., Schroff, F., Belongie, S.,
  Perona, P.: Caltech-ucsd birds 200  (2010)

\bibitem{ref-14}
Wu, P., Zhai, W., Cao, Y.: Background activation suppression for weakly
  supervised object localization. In: Proceedings of the CVPR. pp. 14228--14237
  (2022)

\bibitem{ref-6}
Wu, P., Zhai, W., Cao, Y., Luo, J., Zha, Z.J.: Spatial-aware token for weakly
  supervised object localization. In: Proceedings of the ICCV. pp. 1844--1854
  (2023)

\bibitem{ref-13}
Xie, J., Luo, C., Zhu, X., Jin, Z., Lu, W., Shen, L.: Online refinement of
  low-level feature based activation map for weakly supervised object
  localization. In: Proceedings of the ICCV. pp. 132--141 (2021)

\bibitem{ref-15}
Xie, J., Xiang, J., Chen, J., Hou, X., Zhao, X., Shen, L.: C2am: Contrastive
  learning of class-agnostic activation map for weakly supervised object
  localization and semantic segmentation. In: Proceedings of the CVPR. pp.
  989--998 (2022)

\bibitem{ref-41}
Xu, J., Hou, J., Zhang, Y., Feng, R., Zhao, R.W., Zhang, T., Lu, X., Gao, S.:
  Cream: Weakly supervised object localization via class re-activation mapping.
  In: Proceedings of the CVPR. pp. 9437--9446 (2022)

\bibitem{ref-49}
Xu, L., Ouyang, W., Bennamoun, M., Boussaid, F., Xu, D.: Learning multi-modal
  class-specific tokens for weakly supervised dense object localization. In:
  Proceedings of the CVPR. pp. 19596--19605 (2023)

\bibitem{ref-33}
Xu, M., Yin, X., Qiu, L., Liu, Y., Tong, X., Han, X.: Sampro3d: Locating sam
  prompts in 3d for zero-shot scene segmentation. arXiv preprint
  arXiv:2311.17707  (2023)

\bibitem{ref-12}
Xue, H., Liu, C., Wan, F., Jiao, J., Ji, X., Ye, Q.: Danet: Divergent
  activation for weakly supervised object localization. In: Proceedings of the
  ICCV. pp. 6589--6598 (2019)

\bibitem{ref-27}
Yan, Z., Li, J., Li, X., Zhou, R., Zhang, W., Feng, Y., Diao, W., Fu, K., Sun,
  X.: Ringmo-sam: A foundation model for segment anything in multimodal
  remote-sensing images. IEEE Transactions on Geoscience and Remote Sensing
  \textbf{61},  1--16 (2023)

\bibitem{ref-34}
Yang, J., Gao, M., Li, Z., Gao, S., Wang, F., Zheng, F.: Track anything:
  Segment anything meets videos. arXiv preprint arXiv:2304.11968  (2023)

\bibitem{ref-8}
Yao, Y., Wan, F., Gao, W., Pan, X., Peng, Z., Tian, Q., Ye, Q.: Ts-cam: Token
  semantic coupled attention map for weakly supervised object localization.
  IEEE Transactions on Neural Networks and Learning Systems p. 1–13 (Jan
  2022)

\bibitem{ref-32}
Yu, T., Feng, R., Feng, R., Liu, J., Jin, X., Zeng, W., Chen, Z.: Inpaint
  anything: Segment anything meets image inpainting. arXiv preprint
  arXiv:2304.06790  (2023)

\bibitem{ref-16}
Zhang, C.L., Cao, Y.H., Wu, J.: Rethinking the route towards weakly supervised
  object localization. In: Proceedings of the CVPR. pp. 13460--13469 (2020)

\bibitem{ref-48}
Zhang, D., Han, J., Cheng, G., Yang, M.H.: Weakly supervised object
  localization and detection: A survey. IEEE Transactions on Pattern Analysis
  and Machine Intelligence  \textbf{44}(9),  5866--5885 (2021)

\bibitem{ref-29}
Zhang, X., Liu, Y., Lin, Y., Liao, Q., Li, Y.: Uv-sam: Adapting segment
  anything model for urban village identification. In: Proceeding of the AAAI
  (2024)

\bibitem{ref-1}
Zhou, B., Khosla, A., Lapedriza, A., Oliva, A., Torralba, A.: Learning deep
  features for discriminative localization. In: Processing of the CVPR (2016)

\bibitem{ref-42}
Zhu, L., Chen, Q., Jin, L., You, Y., Lu, Y.: Bagging regional classification
  activation maps for weakly supervised object localization. In: Proceedings of
  the ECCV. pp. 176--192 (2022)

\end{thebibliography}
\end{document}